\documentclass[letterpaper]{article} 
\usepackage[]{aaai25}  
\usepackage{times}  
\usepackage{helvet}  
\usepackage{courier}  
\usepackage[hyphens]{url}  
\usepackage{graphicx} 
\usepackage{natbib}  
\usepackage{caption} 
\usepackage{algorithm}
\usepackage{algorithmic}
\usepackage{newfloat}
\usepackage{listings}
\usepackage{tabularx}
\usepackage{booktabs}
\usepackage[dvipsnames]{xcolor}
\usepackage{enumitem} 
\usepackage{amssymb}
\usepackage{tcolorbox}
\usepackage{colortbl}
\DeclareCaptionStyle{ruled}{labelfont=normalfont,labelsep=colon,strut=off} 
\floatstyle{ruled}
\newfloat{listing}{tb}{lst}{}
\floatname{listing}{Listing}
\title{When Fairness Isn’t Statistical: The Limits of Machine Learning in Evaluating Legal Reasoning}

\author {
    Claire Barale\textsuperscript{\rm 1},
    Michael Rovatsos\textsuperscript{\rm 1},
    Nehal Bhuta\textsuperscript{\rm 2}
}
\affiliations {
    \textsuperscript{\rm 1}School of Informatics, The University of Edinburgh\\
    \textsuperscript{\rm 2}School of Law, The University of Edinburgh\\
    claire.barale@ed.ac.uk
}

\begin{document}

\maketitle

\begin{abstract}
Legal decisions are increasingly evaluated for fairness, consistency, and bias using machine learning (ML) techniques. In high-stakes domains like refugee adjudication, such methods are often applied to detect disparities in outcomes. Yet it remains unclear whether statistical methods can meaningfully assess fairness in legal contexts shaped by discretion, normative complexity, and limited ground truth.

In this paper, we empirically evaluate three common ML approaches (feature-based analysis, semantic clustering, and predictive modeling) on a large, real-world dataset of 59,000+ Canadian refugee decisions (\textsc{AsyLex}). Our experiments show that these methods produce divergent and sometimes contradictory signals, that predictive modeling often depends on contextual and procedural features rather than legal features, and that semantic clustering fails to capture substantive legal reasoning.

We show limitations of statistical fairness evaluation, challenge the assumption that statistical regularity equates to fairness, and argue that current computational approaches fall short of evaluating fairness in legally discretionary domains. We argue that evaluating fairness in law requires methods grounded not only in data, but in legal reasoning and institutional context.

\end{abstract}

\section{Introduction}

Machine learning is increasingly used to evaluate decision-making in high-stakes domains such as healthcare, education, and law \cite{garg2025legalai}. Additionally, in the legal sector, the push toward automation and data-driven evaluation is driven by growing caseloads, backlogs, and resource constraints. As a result, computational tools are increasingly used in both the public sector and the legal sector \cite{engstrom2020government} for their potential to analyze large amounts of historical data, to improve and evaluate consistency and transparency.

However, fairness in legal decision-making is difficult to assess, and efforts to quantify fairness or detect bias face foundational challenges, as fairness cannot be reduced to equal outcomes. In domains such as refugee adjudication, where outcomes are expected to vary based on case-specific reasoning, statistical disparity does not necessarily signal unfairness. Should similar cases always yield similar outcomes? What counts as a similar case? And what differences are justified? Legal fairness involves not only distributive parity (similar outcomes for similar cases), but also procedural and justificatory fairness, i.e., whether decisions are reasoned, consistent in process, and normatively defensible.

This paper evaluates whether existing empirical methods can detect disparities and evaluate fairness in refugee decisions, using a new and publicly available dataset: \textsc{AsyLex} \cite{barale-etal-2023-asylex}: \textbf{Can machine learning (ML) methods meaningfully evaluate fairness in a legal domain where disparities of outcomes may reflect not bias, but legally justified variation?}. 

\textsc{AsyLex} comprises over 59,000 Canadian refugee determinations, each including the full decision text. To our knowledge, this is the first empirical study to use multiple machine learning methods to assess fairness in refugee adjudication using the \textsc{AsyLex} dataset. Additionally, it is the first study to analyze a refugee law corpus that includes full-text legal decisions, rather than relying solely on structured or statistical features.

The question of whether empirical methods can detect legally unsupported disparities in legal outcomes is particularly important in refugee law, where decisions carry life-altering consequences, legal reasoning is often discretionary, and the potential for unjustified variation threatens trust in the system. Refugee protection decisions have significant implications, with around 9.9 million people recognized as either convention refugees or asylum seekers, according to the 2023 United Nations High Commissioner for Refugees annual report\footnote{UNHCR annual report: \url{https://reporting.unhcr.org/global-report-2023-executive-summary}}. Prior studies have documented troubling outcome variability in refugee decisions, with grant rates differing significantly across judges or cities. However, efforts to quantify such disparities often rely on narrow performance metrics, simplified statistical features, or assume that uniformity of outcomes is a measure of fairness, potentially overlooking the legal complexity of the domain.

We take a more holistic approach and aim not only to surface patterns of statistical variation but to evaluate whether these methods support reliable fairness assessments in a domain where disparities may reflect either bias or justified discretion.

Our study presents empirical observations on \textsc{AsyLex} derived from three computational methods commonly used in analysing text data - feature-based analysis, semantic clustering, and predictive modeling. These approaches allow for comparing a comprehensive range of techniques used in computational social science. We take a comprehensive approach to fairness that aligns with the legal domain, which includes both distributive fairness (ensuring that similar cases receive similar outcomes) and procedural fairness (evaluating whether decision-making processes are applied consistently and whether the reasons for decisions align across cases).


We demonstrate that standard machine learning methods, such as clustering, classification, and feature-based statistical analysis, often overlook procedural fairness in contexts where legal reasoning rather than just outcomes should be considered. Our contributions are threefold:
\begin{enumerate}
    \item We analyze refugee decisions in \textsc{AsyLex}, a dataset that has no been studied for fairness, and show how disparities emerge across time, geography, and claimant demographics, while also identifying cases where such variation reflects justified legal discretion.
    \item We evaluate three common ML approaches (clustering, classification, and feature-based analysis) and demonstrate that they often fail to yield meaningful insights into fairness when applied to legal text. 
    \item We argue that without stronger theoretical and domain-specific grounding, computational assessments of fairness risk oversimplify legal reasoning.
\end{enumerate}



We start by providing background and an overview of previous fairness analyses of refugee decisions. Second, we introduce the \textsc{AsyLex}, and present the methodology of the case study. We then present and compare results obtained with each tested method, before discussing limitations of machine learning methods in the evaluation of legal contexts.

\section{Background and Related Work}
\label{sec:background}

\subsection{Refugee Law and Discretionary Justice}
Refugee status determination is a legally grounded but highly discretionary process. Decisions are made following the 1951 Refugee Convention and its 1967 Protocol\footnote{\url{https://www.unhcr.org/media/convention-and-protocol-relating-status-refugees}}, which protect individuals facing a ``well-founded fear of being persecuted for reasons of race, religion, nationality, membership of a particular social group, or political opinion'' (Art. 1A(2)). These determinations hinge on subjective credibility assessments, interpretive judgments, and context-specific legal reasoning.

Fairness objectives in law are grounded in the Rawlsian principle of justice, which refers to equality of opportunity and is defined as impartial treatment, procedural fairness, and equality before the law. Refugee adjudication exemplifies the complexities of fair decision-making evaluation and prior research and data analysis has documented substantial variability in asylum outcomes depending on judges or years, often referred to as \textit{refugee roulette} or the \textit{luck of the draw} \cite{ramji2007refugee, legomsky2007learning, rehaag2012judicial}, a phenomenon that has also gained attention in mainstream media \cite{preston2007asylum}. 

While such variability raises concerns about fairness, Rehaag \cite{rehaag2019judicial} notes that observed outcome differences may reflect legitimate variation in case types, claimant profiles, or judicial specialization. This paper advances this line of work by conducting a detailed empirical analysis of what different machine learning methods reveal and fail to reveal about fairness in refugee adjudication. 

\subsection{Fairness in Machine Learning}

Fairness in decision-making is often defined as the absence of systematic favoritism or disadvantage based on inherent or socially significant attributes \cite{mehrabi2021survey}. In machine learning, fairness is viewed from two main angles: distributive fairness, which assesses equality of outcomes among individuals or groups, and procedural fairness, which emphasizes the justification and consistency of the processes that generate those outcomes \cite{grgic2018beyond, morse2021ends, wang2024procedural}. The machine learning literature largely focuses on distributive fairness metrics. It defines it through formal statistical criteria at the group or individual level \cite{dwork2012fairness}, such as demographic parity, equal opportunity, and equalized odds \cite{hardt2016equality, passi2019problem}. These metrics often presume clean, structured data, established ground truth, and uniform decision rules, assumptions not upheld in discretionary legal systems.

In fairness-aware ML, disparities in outcomes between comparable cases are often used as indicators of potential bias. In this work, disparity is defined as an observable statistical variation in outcomes across cases that appear similar by some measure of similarity that we define (such as the same country of origin).  However, in refugee adjudication, protected group status (race, religion, nationality, membership of a particular social group, or political opinion) is part of the legal grounds for granting protection, complicating the application of standard notions of bias and parity. Thus, a strict application of statistical fairness metrics may incorrectly interpret legitimate legal discretion as unfairness and apply an unsuitable framework to that context. Beyond outcome parity, legal fairness requires attention to the justifications behind decisions and the institutional and contextual constraints surrounding those decisions. In other words, disparities detected by fairness analysis may be fair, and disparity becomes unfairness only when outcome differences cannot be accounted for by legitimate, legally salient factors.

\subsection{Using Machine Learning to Assess Disparities in Outcomes in Legal Contexts}
In addition to statistical analysis of decision \cite{rehaag2019judicial}, previous research has evaluated unequal treatments in legal decision-making using machine learning to predict
legal outcomes across domains such as criminal sentencing \cite{wang2021equality}, Supreme Court decisions \cite{chen2019mood}, and immigration law \cite{chen2017can, piccolo2023predicting}. Categorical features presented as unrelated to the merits of a case, such as the identity of the judge, the time of year, or the hearing location, have been shown to be reliable predictors of outcomes of asylum decisions \cite{chen2017can, dunn_sagun, rehaag2023luck}. 
One of the methods we use (Method 3) is inspired by this line of work. However, we argue that a model that utilizes non-legally salient features and still achieves high accuracy does not necessarily imply that the decision-making process and outcome are unfair.

\subsection{Challenges of Fairness Evaluation in Legal Texts}
Since most legal data consists of unstructured text, we base our analysis on a real-world dataset of complete refugee decision texts. This approach enables us to assess fairness not in a simplified or artificial environment, but in the actual linguistic and institutional complexities of legal decision-making as it happens in practice.

Evaluating fairness from unstructured legal text introduces granularity and precision, but also further challenges, including ambiguity, noise, lack of ground truth, and complex context-dependent reasoning \cite{zhao2023fairness, rogers-2021-changing, keith-stent-2019-modeling, piccolo2023predicting}. Fairness evaluations in natural language processing (NLP) often rely on the availability of clear demographic labels and consistent annotation guidelines. However, these assumptions are rarely met in legal documents, which frequently experience low inter-annotator agreement and contain noisy metadata. Although these issues represent limitations in our study, they also highlight the real-world constraints faced in legal contexts and the challenges that practitioners encounter when trying to assess fairness.



\section{Dataset and Problem Setup} \label{sec:dataset}

\subsection{\textsc{AsyLex} Overview}
\textsc{AsyLex} is a corpus of 59,112 Canadian refugee decisions from 1996 to 2022. These documents are proceedings of the Canadian Refugee Protection Division (RPD) and Refugee Appeal Division (RAD, 92.19\% of the cases). Documents contain narrative descriptions of claims, credibility assessments, legal analyses, and final decisions, reporting the entirety of a case, from initial arguments to the ultimate judgment. Outcomes are binary (grant/reject: refugee status granted for 27.1\%, rejected for 72.9\% of the cases), but the underlying justifications vary widely. All documents are available online and hosted by the Canadian Legal Information Institute (CanLII), and the dataset has previously been anonymized and manually annotated for legally relevant features \cite{barale-etal-2023-asylex}. \textsc{AsyLex} does not include all Canadian refugee decisions from 1996 to 2022, but consists of a sample made up of cases publicly released during that time, while other cases are not released for privacy reasons. This sampling bias comes from how cases are documented in the first place, rather than being an inherent methodological problem. While no larger dataset is currently accessible to researchers, this sample may still over-represent or under-represent certain judges, claimant groups, or regions. Our findings should be viewed as indicative of patterns within this dataset rather than definitive statistics for the entire population of refugee decisions. This perspective does not diminish our assertion regarding the usability of machine learning (ML) methods in evaluating legal fairness, as we compare the methods to one another rather than to a ground truth.

\subsection{Controlled Subset}
To probe how outcome patterns and model behavior persist under reduced variance, we construct a \textsc{controlled subset} of decisions from six judges: three with consistently low grant rates (\textbf{Low-Grant Group Rate, LGR}) and three with consistently high grant rates (\textbf{High-Grant Rate Group, HGR}). These judges were selected based on the number of decisions and grouped by decision outcomes: Judges 146, 465, and 497 (LGR) issued exclusively negative decisions, while Judges 280, 154, and 346 (HGR) had above-average grant rates. This subset contains 592 decisions - 427 rejections and 165 grants.

By collapsing individual judges into LGR and HGR groups, we simulate aggregated judicial behavior at both ends of the grant rate spectrum. To validate our choice of judges, we conducted a two-sample significance test (chi-square). The low resulting p-value (1.76 $\times$ 10$^{-27}$) confirms that the difference in grant rates between groups is statistically significant.

\subsection{Problem Description}
Our goal is to examine whether ML methods can detect similar patterns of disparity in these decisions and whether those patterns meaningfully reflect fairness concerns. We aim to evaluate how effectively each method can both confirm known disparities and independently detect patterns without prior observation. Additionally, we examine whether the methods identify consistent patterns, i.e., whether their findings are reliable and whether one method’s results corroborate another’s.

We focus on four axes of potential disparity, derived from prior literature \cite{rehaag2023luck, ramji2007refugee, dunn_sagun, chen2017can} and legal consultation. Each dimension corresponds to a specific hypothesis, offering a potential explanatory framework to analyze variability in decision outcomes.

\begin{itemize}
    \item \textbf{Temporal}: Do outcomes vary significantly across years? We analyze changes in outcomes and decision-making patterns over time within \textsc{Asylex}, which spans from 1996 to 2022. This allows us to investigate whether judges' decisions are influenced by policy shifts, politics, or other temporal factors and whether some judges preside only during specific periods, potentially aligning their decisions with particular political or institutional contexts. 

    \item \textbf{Geographic}: Are some cities more likely to grant protection than others? Do judges in certain cities exhibit greater leniency? We examine patterns based on the location of hearings by city, which could reflect regional variations, such as differences in local policies.
    
    \item \textbf{Demographic}: Do claimant characteristics (gender, sexual orientation, country of origin) correlate with outcomes? Do judges specialize in cases based on the claimant's gender? Do judges specialize in cases involving minors? Do judges specialize in cases involving LGBTQIA+ claimants? Do some judges specialize in claimants from specific geographic regions? We analyze changes in decision-making patterns based on claimant attributes, a dimension particularly relevant for subsequent fairness analyses.

    \item \textbf{Legal grounds}: Do judges rely differently on legal justifications? Legal grounds are defined as the 5 reasons stated in the Refugee Convention to grant refugee protection: race, religion, nationality, membership of a particular social group, or political opinion.
\end{itemize}

\section{Methodology and Experimental Setup}
\label{sec:methodo}
We employ three classes of methods to explore patterns in the \textsc{AsyLex} dataset, chosen to reflect different analytical choices common in real-world text data analysis and to expose the limits of ML fairness assessments in law. Together, these methods let us ask: Which disparities are detectable by each method? Where do their signals diverge? And what does that say about their reliability for evaluating legal fairness?

\paragraph{Method 1: Feature-Based Analysis}
This method relies on extracted categorical features \cite{barale-etal-2023-asylex} to examine disparities across predefined variables in \textsc{Asylex} (e.g., judge, city, year, demographic attributes). We conduct a novel in-depth data analysis to assess patterns, perform statistical evaluations, and identify trends or disparities. This method is interpretable and hypothesis-driven, but dependent on feature extraction quality.

\paragraph{Method 2: Semantic Clustering via Embeddings}
Using OpenAI’s \textit{text-embedding-3-small}, we embed full decision texts and apply K-means clustering to detect semantically similar cases. We then examine whether clusters correlate with outcomes, judge identity, or legal grounds. Transforming documents into rich vector representation allows us to cluster documents based on semantic similarity without relying on categorical data extraction and investigate whether similar cases lead to consistent or divergent outcomes. 

The distance between any two vectors reflects their semantic relatedness: smaller distances indicate higher similarity, while larger distances suggest lower similarity. Before clustering, the text undergoes preprocessing to remove common words (a custom list of stopwords) and duplicates, ensuring that only the most relevant and differentiating tokens remain for analysis. This method captures linguistic similarity but lacks legal interpretability and explainability of detected patterns.

\paragraph{Method 3: Predictive Modeling}
We train random forests and simple neural classifiers to predict grant/reject decisions using either structured features or text. Predictive modelling is particularly valuable in contexts where ground truth labels for decisions are unavailable.  

We define the problem as a binary classification task and classify 24 years of refugee proceedings, using first the features (random forest) and second the free text (simple feedforward neural network). These models serve as proxies to reflect the decision-making process, highlight factors that increase the likelihood of an application being accepted, and identify patterns, disparities, or learned biases that may be present in the classifier's predictions.



The Random Forest is chosen for its interpretability, efficiency, and widespread use in legal prediction tasks \cite{chen2017can, dunn_sagun}. Its built-in feature importance measures allow us to analyze which variables drive outcome predictions. We perform hyperparameter tuning using stratified 5-fold cross-validation and oversample the minority class (rejections) to account for imbalance. To complement this, we train a fully connected neural network on full decision text. The architecture is minimal (two hidden layers with ReLU activations) and serves as a baseline for evaluating how much textual signal is available for outcome prediction.

\paragraph{\textbf{Feature Importance}}
Feature importance in the Random Forest is computed via impurity reduction measured using the Gini index, averaged across all decision trees. This allows us to assess whether predictive success is driven by legally grounded signals (e.g., legal grounds) or procedural artifacts (e.g., year, judge ID).


\paragraph{\textbf{Evaluation Metrics}}
We report macro-averaged F1 score, ROC AUC, and test-set Accuracy averaged over 5 runs. To assess fairness, we apply the \textit{Equality of Opportunity} (EO) criterion \cite{hardt2016equality}, a relaxed version of equalized odds, which requires equal true positive rates (recall) across protected groups. EO was chosen for its alignment with legal principles, its common use in the fairness literature, its focus on favourable outcomes, and its empirical relevance. EO is assessed by comparing recall, measuring the model's ability to identify positive outcomes across groups. Any variation in Recall indicates potential disparities. EO requires an equal chance of success across groups, i.e., non-discrimination within the outcome group granted with success ($Y=1$). 


\section{Results and Evaluation}
\label{sec:results}
We report results from the three methods described in the Methodology Section above, and compare the findings of each method against those of the other two, assessing the consistency and complementarity of their insights. For each method, we summarize key findings organized by dimension, explicitly state the limitations, and compare them to the other methods. While all methods detect disparities, none reliably distinguish between normatively troubling disparities and legitimate variation rooted in legal reasoning. 

\begin{figure}[t]
\centering
\includegraphics[width=0.9\columnwidth]{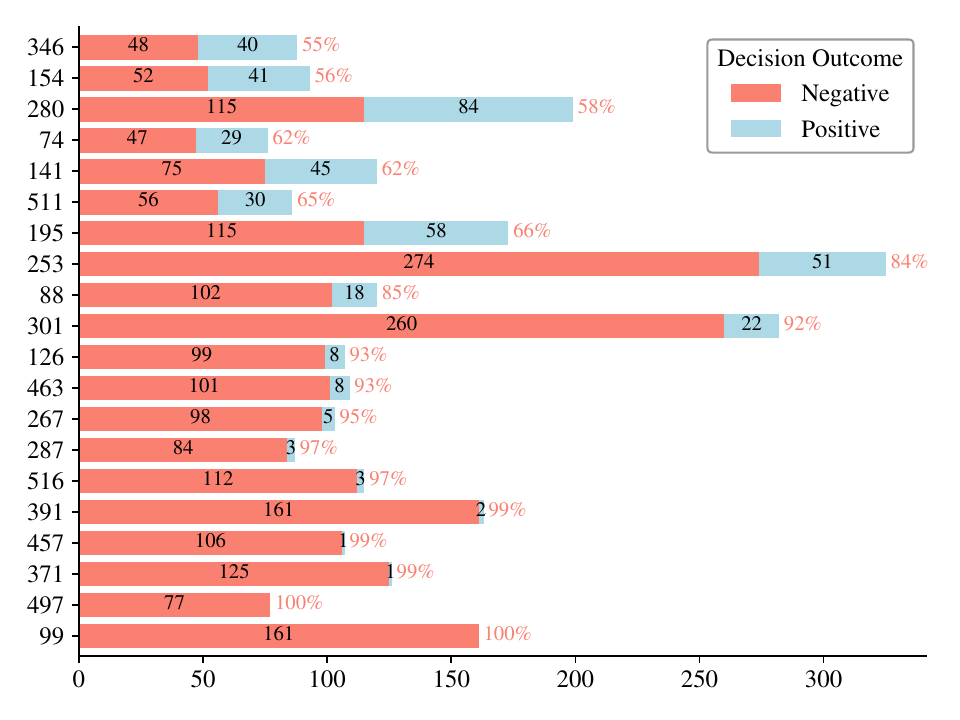} 
\caption{Decision outcomes for the 25 judges who rendered the highest number of decisions. Each bar represents the count of decisions
per judge (x-axis), with the percentage of negative decisions to the right of each bar. The y-axis shows the judge identifiers.}
\label{fig1:decision_outcomes}
\end{figure}

\subsection{Feature-Based Statistical Analysis (Method 1)}
\label{sec:method1-results}
\begin{figure}[t]
\centering
\includegraphics[width=0.9\columnwidth]{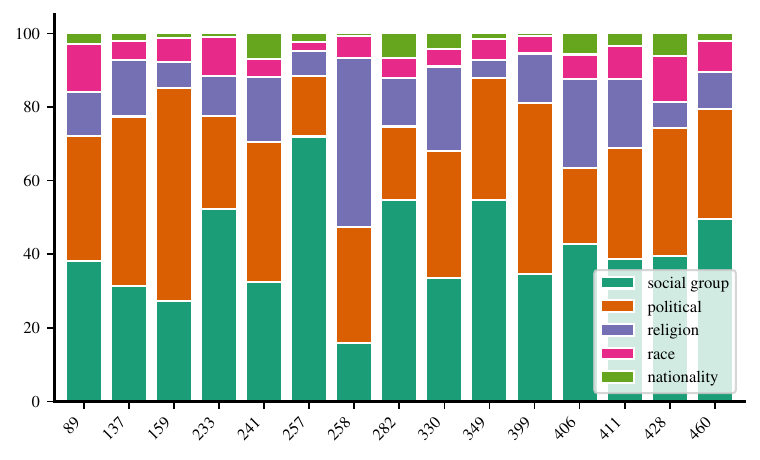} 
\caption{Distribution of legal grounds (y-axis, normalized frequency) cited by the 15 judges (x-axis) who rendered the most decisions in the \textsc{AsyLex} dataset.}
\label{fig2:legalgrounds}
\end{figure}

\begin{tcolorbox}[colback=blue!2!white,colframe=blue!30!black, boxrule=0.4pt, arc=2pt]
\footnotesize
Significant grant rate disparities appear across time (up to 37.6\%), geography (28.4\%), judge identity (0–100\%), legal grounds cited (Fig.\ref{fig2:legalgrounds}), and claimant characteristics. 
\end{tcolorbox}

\paragraph{Judge-Level Disparities}
Grant rates among judges range from 0\% to 100\% (Fig. \ref{fig1:decision_outcomes}). This variation aligns with previous literature findings on Canada asylum adjudication, although evaluated on a different set of cases \cite{rehaag2012judicial}. 

\paragraph{Temporal}
Grant rates fluctuate sharply year to year, showing a 37.6\% swing between 2000 (2.5\%) and 2012 (40.1\%). While these shifts could reflect geopolitical events or legal reforms, they also suggest instability in how claims are evaluated across time.

\paragraph{Geographic}
Outcome variation by city of hearing is substantial. For instance, the grant rate in Ottawa is 28.4\% lower than in Winnipeg. Differences persist even when controlling for the claimants' top countries of origin. Such geographic disparity may stem from institutional norms or scheduling logistics, but the scale of variation warrants scrutiny.

\paragraph{Demographic}
Claimant characteristics correlate with significant outcome differences, compared to an average grant rate of 26.3\%:
\begin{itemize}
    \item Female claimants: 20\% grant rate (-6.3\%);
    \item LGBTQIA+ claimants: 12.2\% grant rate (-14.1\%);
    \item Minors: 16.4\% grant rate (-9.9\%).
\end{itemize}

Moreover, our study on the \textsc{controlled subset}  shows these groups are disproportionately assigned to the LGR (Low Grant Rate) judges. However, when these cases are heard by HGR (High Grant Rate) judges, outcomes improve substantially, +20\% for minors, +54\% for LGBTQIA+ claimants. This suggests that cases are not randomly assigned.

\paragraph{Legal Grounds}
Fig. \ref{fig2:legalgrounds} reports the legal ground cited by the 15 judges who rendered most decisions. Further analysis shows that judges in the LGR group more frequently cite religion (in 43.2\% of cases) or nationality (37.09\%), while HGR judges invoke political opinion (24.3\%) and race (37.17\%). Critically, grant rates differ drastically by legal ground: LGR judges grant 0\% when citing nationality, while HGR judges grant 60\% when citing political opinion, a 49.9\% gap. These differences suggest varying outcomes and divergent interpretations.

\paragraph{Limitations}
Feature-based statistical analysis identifies measurable disparities across variables like judges, cities, and claimant groups. However, it does not explain the origins of these disparities or assess whether they are problematic. Without causal modeling, it reveals correlations without addressing underlying reasons. For example, variations in outcomes among judges might result from different adjudicative philosophies or procedural specializations, such as consistently handling cases from certain countries. Thus, statistical disparities cannot be equated with unfairness without understanding their mechanisms.

This method also has limitations due to data availability and design choices. We rely on predefined features, which can introduce selection bias and restrict discovery to known disparities. It overlooks potential interactions between features and may be affected by noise in feature extraction or incomplete metadata, undermining reliability. 

This method does not scale well with increased feature complexity and is best suited for exploratory analysis, such as searching for hypotheses, and does not allow for exploring legal reasoning found in text.

\subsection{Semantic Clustering (Method 2)}
\label{sec:method2-results}

\begin{tcolorbox}[colback=blue!2!white,colframe=blue!30!black, boxrule=0.4pt, arc=2pt]
\footnotesize
Semantic clustering reveals clear linguistic divergence across decisions, particularly between high- and low-grant rate judges and outcomes. However, these clusters do not align consistently with legal justifications, temporal trends, or protected attributes, highlighting a gap between linguistic similarity and legal reasons.
\end{tcolorbox}

\begin{figure}[t]
\centering
\includegraphics[width=0.8\columnwidth]{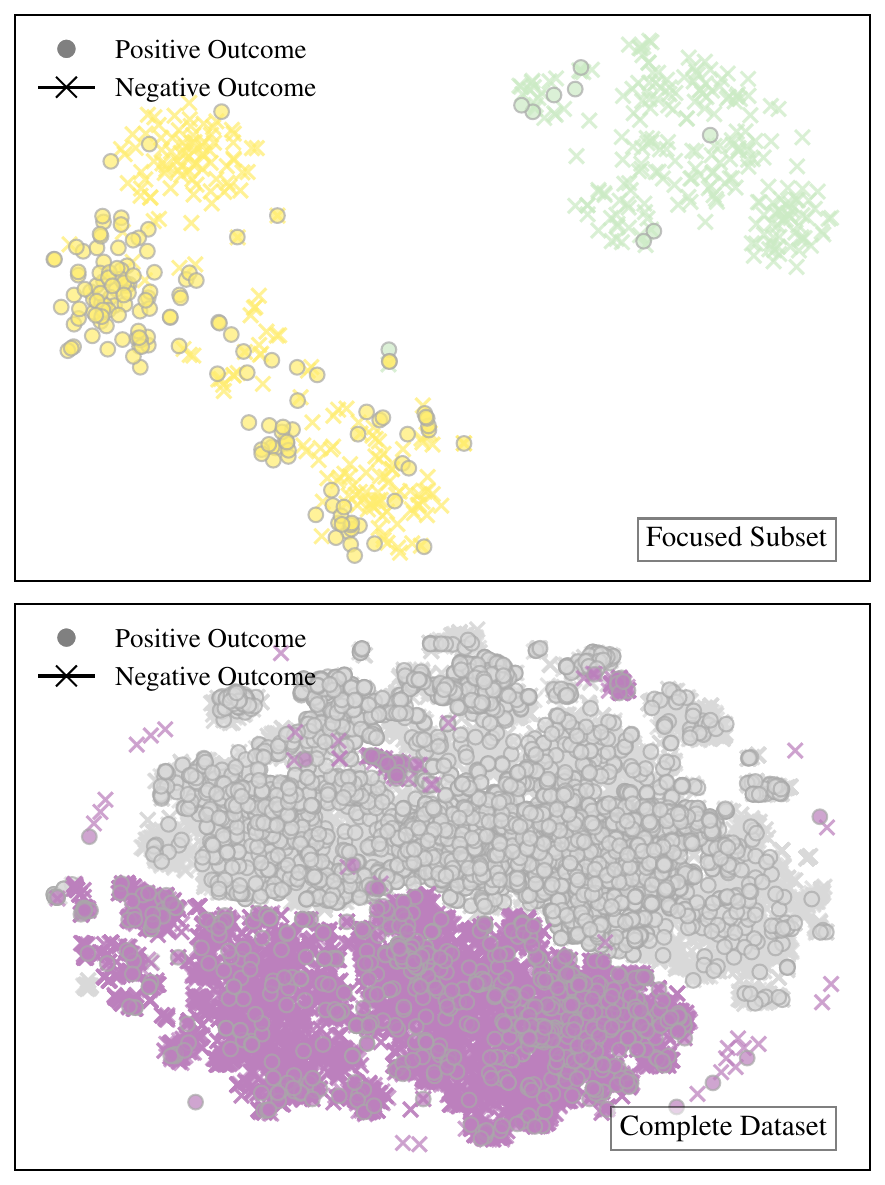} 
\caption{Visualization of clusters and outcomes using t-SNE \cite{tsnevan2008visualizing} for the \textsc{controlled subset} and the full dataset. Outcomes align with cluster split.}
\label{fig3:clusters}
\end{figure}

We identify two major clusters using K-means on the embedded representations of both the full dataset (\textsc{complete dataset}) and the \textsc{controlled subset}. As shown in Fig. \ref{fig3:clusters}, these clusters show a clear split and partial alignment with case outcomes.

\paragraph{Outcome Separation and Judge Correlation}
Clustering shows a marked correlation with case outcomes and judge identity. 
In the \textsc{controlled subset}, \textsc{Cluster 1} is dominated by LGR judges (85.8\%), while \textsc{Cluster 2} is composed almost entirely of HGR judges (98.1\%). This suggests a variation in the language used by each group. On the \textsc{complete dataset} language used in each cluster reflects this split: \textsc{Cluster 1} contains just 7.2\% positive outcomes, while \textsc{Cluster 2} includes 40.6\%, capturing 89.3\% of all granted decisions in \textsc{Asylex}. This suggests that linguistic patterns learned by the model correlate with decision leniency. 

\paragraph{Temporal and Demographic Inconclusiveness}
Clustering does not capture variation along several important fairness dimensions. We observe no significant differences in the year of decision, hearing city, gender of claimant, cases involving minors, or LGBTQIA+ references. Those features yield high p-values ($> 0.05$, chi-square test), suggesting that temporal and demographic features do not structure the clusters, despite being recognized as important factors of outcome variation by Method 1. This raises concerns regarding the method's ability to identify patterns relevant to fairness.

\paragraph{Citizenship Patterns: Surface Signal}
Clusters also show alignment with claimants’ countries of citizenship. For instance, in the \textsc{controlled subset}, Chinese and Indian claimants, two countries of origin with the highest number of claims in \textsc{Asylex}, appear overwhelmingly in \textsc{Cluster 2}. In the \textsc{complete dataset}, countries like Nigeria and Haiti skew heavily toward a single cluster (+9 and +13\% respectively). These patterns raise concerns about group-based treatment but offer limited insight into whether decisions were fair or legally justified

\paragraph{Legal Grounds: Partial Alignment}
Alignment between legal grounds and clusters is partial. In the \textsc{controlled subset}, \textsc{Cluster 1} is more likely to cite social or political group persecution, while \textsc{Cluster 2} more often references race, showing a 33.3\% difference in race as a cited legal ground. Thus, observations on \textsc{Cluster 1} do not align with observations made with Method 1. In the \textsc{complete dataset}, \textsc{Cluster 1} cites social and political groups more often, while \textsc{Cluster 2} leans toward religion and social group justifications. 

Though these differences are statistically significant (p-value $< 10^{-4}$), their interpretability is limited. Further keyword analysis reveals that \textsc{Cluster 1} tends to emphasize credibility issues, while \textsc{Cluster 2} contains more references to family ties and vulnerability. This aligns with the structure of refugee law, where negative decisions are often justified by a lack of credibility, resulting in more uniform language. In contrast, positive decisions may reflect a wider range of justifications, leading to greater linguistic variability.

\paragraph{Limitations}
Semantic clustering organizes cases based on linguistic similarity rather than legal reasoning. It assumes that cases with similar language reflect comparable content or outcomes. However, in legal contexts, particularly refugee adjudication, this assumption often fails. Substantive legal reasoning is usually found only in the final paragraphs of decisions, leading clustering to capture surface-level patterns instead of legally significant justifications. This misalignment can obscure meaningful legal categories identified through feature-based analysis.

Unlike Method 1, clustering does not require predefined features or explicit hypotheses, allowing for exploratory analysis of free text. However, this comes with interpretability challenges. While some alignment between clusters and judge behavior or case outcomes was observed, validating or invalidating hypotheses for some features of interest (gender, minors, LGBTQIA+ status, or year and location of decision) has been inconclusive due to not being statistically significant.

Moreover, semantic clustering is sensitive to the choice of embedding models, clustering algorithms, and hyperparameters. High-dimensional representation learning can lead to patterns that are artifacts rather than legally relevant, complicating interpretation.

In summary, while clustering enables unsupervised discovery in text data, it struggles to differentiate legally justified variations from incidental similarities and offers limited support for fairness evaluations based on legal reasoning.

\subsection{Predictive Modeling (Method 3)}
\label{sec:method3-results}

\begin{tcolorbox}[colback=blue!2!white,colframe=blue!30!black, boxrule=0.4pt, arc=2pt]
\footnotesize
Outcome prediction can achieve high accuracy, reaching up to 93.8\%. However, models primarily depend on procedural and contextual features while legally significant justifications and claimant attributes play only a minor role. Although group-wise recall varies, fairness metrics do not allow for considering justification for the decisions.
\end{tcolorbox}

\paragraph{Model Performance and Feature Combinations}
We trained random forest classifiers using incremental subsets of structured features, described below, on both the \textsc{complete dataset} and the \textsc{controlled subset}. Table \ref{table:tab1_results_subsetoffeatures} reports performance across feature combinations using macro F1, ROC AUC, and Accuracy.

\begin{description}[left=0pt]
    \item[(Part 1) Early Predictability --] Contains four features that take into account only the data available up to the date of the trial, as if a judge was presented with a new case: claimant attributes (gender, origin, age), locations, and documentary evidence (e.g., passports, letters).
    \item[(Part 2) External Features --] Includes external case-related factors: dates, city, tribunal, hearing type (public/private, virtual/in-person), judge ID, and final decision date.
    \item[(Part 3 ) Explanations --] Captures decision justifications: legal ground (race, religion, nationality, social group, political opinion), credibility, and other explanations.
    \item[(Part 4) Judge --] Focuses solely on the judge identifier.
    \item[(Part 5) Claimant Information --] Includes claimant details: citizenship, minor or not, gender, and sexual orientation if specified.
    \item[(Part 6) Locations --] Tests the predictive power of two location features (nouns and adjectives).
\end{description}

\begin{figure}[t]
\centering
\includegraphics[width=0.9\columnwidth]{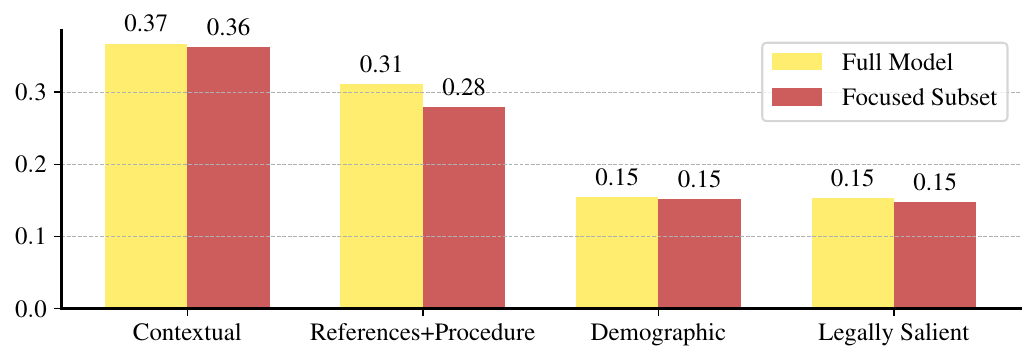}
\caption{Aggregated feature importance for the full model (random forest).
Contextual features dominate the predictions. In contrast, legal reasons features make a smaller contribution, which limits the model's utility in terms of fairness. Features are grouped into four categories: (1) Contextual —  judge, dates, tribunal, date decision, city hearing, organization, public/private, and virtual; (2) References+Procedure — legal citations, elements of procedure, and evidence supporting the case; (3) Demographic — features describing claimant attributes and country of origin;(4) Legally Salient — features central to legal reasoning such as credibility, legal ground, and explanation.}
\label{fig4:featureimp}
\end{figure}

Contrary to the disparities observed with Method 1, judge identity (Part 4) does not strongly predict outcomes in the predictive models. In the \textsc{complete dataset}, explanations (Part 6) yield the highest macro F1 and accuracy scores, suggesting that they contribute to the learning process. However, their performance drops in the \textsc{controlled subset}  and still falls short of the Full Model’s performance by 7.7\% . In contrast, external features (Part 2) achieve higher scores in the \textsc{controlled subset} , indicating that these contextual signals play a greater role in smaller, more homogeneous samples with a clear decision boundary.

\begin{table}[!ht]
    \centering
\begin{tabular}{lccc}
\midrule
Feature space& $\mu$ F1 & Roc Auc & Acc. (test) \\
\midrule

\multicolumn{4}{c}{\textbf{ \textsc{Complete dataset} }}\\
(Part 1) Early pred    &  80.9 &  89.8 &  73.9 \\
        (Part 2) External      &  80.7 & 91.9 &  73.4 \\
        (Part 3) Locations     &  79.5 & 90.5 &  73.5 \\
        (Part 4) Judge         &  61.6 &  67.6 &  73.2 \\
        (Part 5) Claimant info &  76.7 & 88.2 &  73.0 \\
        (Part 6) Explanations  &  82.1 & 90.9 &  74.1 \\
       \midrule 
        \textbf{Full Model}    & \textbf{89.8} & \textbf{97.1} & \textbf{79.1} \\

\midrule
\midrule
\multicolumn{4}{c}{\textbf{ \textsc{Controlled Subset} }} \\
(Part 1) Early pred &  81.5 \textcolor{YellowGreen}{\(\uparrow\)}  & 94.4 \textcolor{YellowGreen}{\(\uparrow\)} &74.4 \textcolor{YellowGreen}{\(\uparrow\)}\\
(Part 2) External  & 87.4 \textcolor{YellowGreen}{\(\uparrow\)}  & 95.7 \textcolor{YellowGreen}{\(\uparrow\)}& 79.0 \textcolor{YellowGreen}{\(\uparrow\)}\\
(Part 3) Locations & 83.7 \textcolor{YellowGreen}{\(\uparrow\)}& 93.0\textcolor{YellowGreen}{\(\uparrow\)} & 75.0 \textcolor{YellowGreen}{\(\uparrow\)}\\
(Part 4) Judge& 70.3 \textcolor{YellowGreen}{\(\uparrow\)}&75.0 \textcolor{YellowGreen}{\(\uparrow\)}& 70.0\textcolor{Mahogany}{\(\downarrow\)} \\
(Part 5) Claimant info & 78.4 \textcolor{YellowGreen}{\(\uparrow\)}& 96.9 \textcolor{YellowGreen}{\(\uparrow\)}& 72.4\textcolor{Mahogany}{\(\downarrow\)} \\
(Part 6) Explanations & 83.3\textcolor{Mahogany}{\(\downarrow\)} & 95.1 \textcolor{YellowGreen}{\(\uparrow\)}& 72.0 \textcolor{Mahogany}{\(\downarrow\)}\\
\midrule 
\textbf{Full Model} & 93.8 \textcolor{YellowGreen}{\(\uparrow\)}&  98.9\textcolor{YellowGreen}{\(\uparrow\)}&  77.0\textcolor{Mahogany}{\(\downarrow\)} \\
\bottomrule
\end{tabular}
\caption{Performance across feature subsets for the \textsc{complete dataset} (upper) and the \textsc{controlled subset} (lower). Green arrows (\textcolor{YellowGreen}{\(\uparrow\)}) indicate an increase in performance compared to the \textsc{complete dataset}, and red arrows (\textcolor{Mahogany}{\(\downarrow\)}) indicate a decrease. Models using procedural features outperform those using substantive legal or demographic information.}\label{table:tab1_results_subsetoffeatures} 
\end{table}

\paragraph{Which Features Matter?}
Fig. \ref{fig4:featureimp} displays aggregated feature importance scores for the full model. Contextual and procedural features dominate, while legally salient features contribute relatively little. 

\paragraph{Temporal} Feature importance analysis shows that the most influential predictors are temporal and procedural features. The hearing date ranks as the top feature across both datasets, indicating that temporal factors significantly drive the model's accuracy. Further study of models accuracy per year reveals that outcome prediction scores vary by up to 5\% across years. Thus, observations differ from those of Method 2, but align with Method 1. 

\paragraph{Geographic} Geographic indicators, such as the city where the hearing takes place, play a smaller role and do not rank highly in predictive power, which contradicts observations made with Method 1.

\paragraph{Demographic} Demographic features, including the claimant's gender, age, and LGBTQIA+ status, collectively contribute only 5.6\% to feature importance, and reach relatively low accuracy in terms of prediction (Table \ref{table:tab1_results_subsetoffeatures}). This differs from observations made with Method 1 that show disparities in outcomes across groups, and aligns with Method 2 that shows no significance of demographic claimants' attributes for clustering decisions. This raises concerns about whether the model effectively captures individual-level variations that are important for fairness.

\paragraph{Legal Grounds} Legal justifications, such as assessments of credibility and grounds for persecution, consistently rank below procedural cues. While features providing written explanations achieve a relatively high accuracy of 74.1\% and rank third in overall feature importance, they still fall short compared to context-based signals like the date of the hearing and the identity of the judge (Fig. \ref{fig4:featureimp}). This limitation restricts the model’s ability to reflect legal reasoning, and those observations differ from those made with Methods 1 and 2, which shows partial alignment.

\paragraph{Group Fairness and Equality of Opportunity}
To evaluate fairness among different demographic groups, we present Recall, AUC, and Accuracy for each group in Table \ref{tab:tab2_recall_groups}. Recall varies between groups, with a performance gap of up to 13.1\% observed in the random forest model. The random forest model predicts grant approvals more accurately for male claimants than for female claimants, showing a difference of +5.3\%.

In contrast to the variability in predictions of the random forest, the neural networks demonstrate greater consistency across groups and achieve a higher overall accuracy (+15.8\%). However, they provide less interpretability into which features drive the prediction, highlighting a trade-off between accuracy and transparency. 

\begin{table}[ht]
\centering
\begin{tabular}{lcc|cc|cc}
\toprule
\textbf{Group} & \multicolumn{2}{c|}{\textbf{Recall}} & \multicolumn{2}{c|}{\textbf{ROC AUC}} & \multicolumn{2}{c}{\textbf{Acc. (test)}}\\
 & RF & NN & RF & NN & RF & NN  \\
\midrule
\textbf{Women} & 80.8 & 99.8 & 95.9 & 99.8 & 75.5 & 89.6  \\
\textbf{Men} & 86.1 & 97.8 & 97.7 & 98.9 & 80.6 & 92.4  \\
\textbf{w/ minors} & 88.7 & 99.5 & 99.2 & 99.7 & 84.9 & 95.5  \\
\textbf{LGBTQIA+} & 93.9 & 98.1 & 91.0 & 98.2 & 88.6 & 94.3  \\
\midrule
\textbf{Full model} & 82.7 & 99.8 & 97.1 & 99.9 & 79.1 & 94.9 \\
\bottomrule
\end{tabular}
\caption{Group-wise evaluation metrics for the \textsc{complete dataset}. We compare results using the random forest (RF) model and our fully connected neural network (NN) model, showing that recall varies across demographic groups.}\label{tab:tab2_recall_groups}
\end{table}

\paragraph{Limitations}
Predictive modeling evaluates the extent to which case outcomes can be predicted based on structured features or free-text representations. The high performance of these models is largely driven by features that correlate with institutional patterns (such as who, when, and where) rather than reflecting legal justification or the individual attributes of claimants. Legal explanations, including credibility assessments and grounds for persecution, consistently rank lower in predictive importance. The model learns to replicate decision-making without needing an initial hypothesis, allowing us to quantify the impact of specific features using a traditional interpretable machine learning approach. However, this process is focused on decision replication rather than fairness evaluation.

Although we report fairness metrics such as group-level recall, these measures are difficult to interpret in the absence of ground truth regarding the correctness or justifiability of decisions. Disparities in model performance across groups may reflect structural biases, but could also result from variations in case composition, incomplete features, or sample imbalance. 

A further limitation of this approach is the potential non-independence of the features used. Finally, the results are inconsistent with the observations made with the two other methods. Variables such as judge ID, location, and claimant origin may be confounded in practice, which complicates attribution of predictive importance and may obscure underlying causes of variation.

Moreover, the results obtained from predictive modeling often diverge from those produced by feature-based analysis or clustering. This inconsistency highlights the method's sensitivity to model assumptions, feature encoding, and data preparation, and suggests that predictive accuracy alone is insufficient for evaluating fairness in legally discretionary contexts.


\subsection{Synthesis: Divergent Signals Across Methods}

\begin{table*}[ht]
\centering
\begin{tabular}{lccc}
\toprule
\textbf{Evaluation Dimension} & \textbf{Feature-Based} & \textbf{Clustering} & \textbf{Predictive Modeling} \\
\midrule
\multicolumn{4}{l}{\textit{Disparity Detection}} \\
\quad Temporal             & \checkmark       & \textbf{X}          & \checkmark        \\
\quad Geographic           & \checkmark       & \textbf{X}          & \textbf{X}  \\
\quad Demographic          & \checkmark       & \checkmark (country only)          & \textbf{X}          \\
\quad Judge      & \checkmark       & \checkmark      & \textbf{X}    \\
\quad Legal Ground  & \checkmark         & \checkmark         & \checkmark           \\
\midrule
\multicolumn{4}{l}{\textit{Legal Reasoning Access}} \\
\quad Uses Full Text       & \textbf{X}           & \checkmark      & \checkmark (NN)       \\
\quad Captures Justification & \textbf{X}         & \textbf{X}          & \textbf{X}            \\
\quad Interpretable Output & high             & low             & high with RF, low with NN           \\
\midrule
\multicolumn{4}{l}{\textit{Fairness Assessment}} \\
\quad Causal Insight       & \textbf{X}           & \textbf{X}          & \textbf{X}            \\
\quad Group Fairness Metric & \textbf{X}          & \textbf{X}          & \checkmark           \\
\quad Consistency Across Methods & low        & low             & low               \\
\bottomrule
\end{tabular}
\caption{Comparison of the three methods. A \checkmark\ indicates that the method supports analysis on that dimension and reveals observable disparities. A \textbf{X} indicates that the method either does not support analysis of that dimension or fails to detect significant disparities. The table also evaluates each method’s ability to engage with legal reasoning, interpretability, and fairness assessment. RF = Random Forest; NN = Neural Network.}
\label{tab3:method-comparison}
\end{table*}

Our analysis reveals that the three methods (feature-based statistical analysis, semantic clustering, and predictive modeling) often yield differing results, summarised in Table \ref{tab3:method-comparison}. For instance, both clustering and predictive modeling yield inconclusive results regarding geographic consistency. While method 2 (clustering) fails to provide conclusive evidence of temporal inconsistencies, feature-based analysis and predictive modeling suggest that the year of the decision impacts the final outcome. Clustering reveals inconsistencies in legal grounds, while predictive modelling remains inconclusive. The divergence in results across methods underscores the absence of a reliable, consistent evaluative signal. Each approach reveals different patterns and draws attention to different disparities, but none can establish causality or offer a robust account of legal reasoning.

Common findings across methods include disparities in claimant assignments, the influence of external features, and variations in grant rates, highlighting that current approaches overwhelmingly focus on outcomes rather than the underlying legal reasoning. Indeed, we find that legally salient justifications, such as the cited ground for protection, were weak signals. Meanwhile, legally secondary or procedural attributes (e.g., date, city) were often influential. This inversion raises concerns about the validity of ML fairness assessments in legal domains. Rather than interrogating the basis of legal decisions, models often reflect institutional artifacts.

Feature-based analysis reveals known disparities but cannot explain their causes. Clustering highlights surface-level divisions rather than patterns of legal reasoning. Predictive models achieve high accuracy in predictions, but they may capture non-legally relevant patterns.

\textbf{In short, patterns identified by one method are not always confirmed by the others}, and drawing definitive conclusions and entirely excluding alternative explanations for these disparities remains challenging due to each method's highlighted limitations. This inconsistency raises concerns about the reliability and interpretability of each method when applied in isolation.

\section{Discussion} \label{sec:discussion}

Can machine learning methods reliably assess fairness as defined within legal frameworks? Our findings suggest that current machine learning (ML) methods are fundamentally limited in this task. None of the methods we tested can reliably evaluate legal fairness because they fail to capture legal reasoning and justification. These tools can surface statistical patterns, but they cannot distinguish between unjust disparities and justified discretion grounded in law.

\paragraph{ML Methods Lack Causal Insight}
A key limitation shared by all these approaches is their inability to establish causality. While they can detect correlations between features and outcomes, such as the relationship between judge identity and grant rates, they cannot determine whether these relationships reflect bias, justified discretion, or unobserved confounders. 

While we report group-level disparities in recall and feature importances, these metrics are correlational and not causal. In refugee law, fairness requires understanding whether group membership unjustly caused an outcome difference. Fairness metrics risk misinterpreting legitimate variation or failing to detect systemic bias without causal modeling, such as counterfactual evaluation. However, applying causal modeling to text, which would be required to grasp the entirety of the reasoning embedded in a case, is challenging because textual data is high-dimensional, unstructured, and lacks explicit mappings between linguistic features and causal variables.

\paragraph{Legal Reasoning}
Similarly, none of the methods can effectively assess the legal reasoning or justification behind a decision, as they lack insight into the underlying deliberative processes and often treat decision text as unstructured input or as simplified and potentially noisy features.

Refugee law exemplifies the complexity of fairness evaluation in law. The legal framework permits and sometimes requires outcome variation based on case, specific facts and subjective credibility assessments. Disparities in grant rates across judges, cities, or time may reflect systemic problems, but they may also result from lawful discretion or unobserved differences in case merit. Our analysis shows that computational methods often cannot tell the difference.

Each method we applied carries implicit assumptions about fairness:
\begin{enumerate}
    \item Feature-based analysis assumes that similar inputs should lead to similar outputs, but overlooks latent variables like claim complexity or evolving legal standards;
    \item Semantic clustering assumes that linguistic similarity implies legal similarity, yet legal meaning often hinges on subtle distinctions in justification or context;
    \item Predictive modeling implies that outcome predictability reflects regularity, when in fact it may capture non legally salient features like hearing date.
\end{enumerate}

Only two of the three approaches (semantic clustering and neural network classification) analyze the decision text directly. However, even these struggle to capture substantive legal reasoning. Much of the legal content, including the nuances of justification and interpretive reasoning, remains inaccessible to the feature-based methods. As a result, the patterns these models uncover tend to reflect procedural artifacts rather than substantive legal distinctions.

\paragraph{Rethinking Fairness Evaluation in Legal Contexts}
This paper challenges common assumptions in prior work on fairness in refugee proceedings that equate statistical disparity with unfairness and rely on predictive accuracy or existing fairness metrics to evaluate legal systems. These metrics may be meaningful in constrained or synthetic settings, but when applied to real-world legal decisions, as shown by our work, they risk oversimplifying a complex process.

In future work, we advocate for a shift away from outcome-based fairness metrics toward approaches that incorporate procedural fairness in the sense of legal reasoning and justification structure. Our study shows a need for designing specific methods, frameworks, and metrics aligned with the legal context that do not solely focus on outcome parity. For instance, future datasets could include text-level annotations of justifications, enabling fairness evaluations that align with legal standards. 

\section{Conclusion}
This paper examined whether standard machine learning methods can meaningfully evaluate fairness in legal decision-making, using a large real-world dataset of refugee determinations. Across feature-based analysis, semantic clustering, and predictive modeling, we found that methods often diverge in their observations. These approaches frequently detect disparities but fail to distinguish between unjust bias and legitimate legal discretion. Crucially, none is capable of evaluating procedural fairness and providing insights into legal reasoning.

\bibliography{custom}

\begin{thebibliography}{25}
\providecommand{\natexlab}[1]{#1}

\bibitem[{Barale et~al.(2023)Barale, Klaisoongnoen, Minervini, Rovatsos, and Bhuta}]{barale-etal-2023-asylex}
Barale, C.; Klaisoongnoen, M.; Minervini, P.; Rovatsos, M.; and Bhuta, N. 2023.
\newblock {A}sy{L}ex: A Dataset for Legal Language Processing of Refugee Claims.
\newblock In \emph{Proceedings of the Natural Legal Language Processing Workshop 2023}, 244--257. Singapore: Association for Computational Linguistics.

\bibitem[{Chen and Eagel(2017)}]{chen2017can}
Chen, D.~L.; and Eagel, J. 2017.
\newblock Can machine learning help predict the outcome of asylum adjudications?
\newblock In \emph{Proceedings of the 16th edition of the International Conference on Articial Intelligence and Law}, 237--240.

\bibitem[{Chen and Loecher(2019)}]{chen2019mood}
Chen, D.~L.; and Loecher, M. 2019.
\newblock Mood and the malleability of moral reasoning.
\newblock \emph{Available at SSRN 2740485}.

\bibitem[{Dunn et~al.(2017)Dunn, Sagun, \c{S}irin, and Chen}]{dunn_sagun}
Dunn, M.; Sagun, L.; \c{S}irin, H.; and Chen, D. 2017.
\newblock Early predictability of asylum court decisions.
\newblock In \emph{Proceedings of the 16th Edition of the International Conference on Articial Intelligence and Law}, ICAIL '17, 233–236. New York, NY, USA: Association for Computing Machinery.
\newblock ISBN 9781450348911.

\bibitem[{Dwork et~al.(2012)Dwork, Hardt, Pitassi, Reingold, and Zemel}]{dwork2012fairness}
Dwork, C.; Hardt, M.; Pitassi, T.; Reingold, O.; and Zemel, R. 2012.
\newblock Fairness through awareness.
\newblock In \emph{Proceedings of the 3rd innovations in theoretical computer science conference}, 214--226.

\bibitem[{Engstrom et~al.(2020)Engstrom, Ho, Sharkey, and Cu{\'e}llar}]{engstrom2020government}
Engstrom, D.~F.; Ho, D.~E.; Sharkey, C.~M.; and Cu{\'e}llar, M.-F. 2020.
\newblock Government by algorithm: Artificial intelligence in federal administrative agencies.
\newblock \emph{NYU School of Law, Public Law Research Paper}, (20-54).

\bibitem[{Garg and Ma(2025)}]{garg2025legalai}
Garg, A.; and Ma, M. 2025.
\newblock Opportunities and Challenges in Legal AI.
\newblock White paper, Stanford Law School.
\newblock CodeX.

\bibitem[{Grgi{\'c}-Hla{\v{c}}a et~al.(2018)Grgi{\'c}-Hla{\v{c}}a, Zafar, Gummadi, and Weller}]{grgic2018beyond}
Grgi{\'c}-Hla{\v{c}}a, N.; Zafar, M.~B.; Gummadi, K.~P.; and Weller, A. 2018.
\newblock Beyond distributive fairness in algorithmic decision making: Feature selection for procedurally fair learning.
\newblock In \emph{Proceedings of the AAAI conference on artificial intelligence}, volume~32.

\bibitem[{Hardt, Price, and Srebro(2016)}]{hardt2016equality}
Hardt, M.; Price, E.; and Srebro, N. 2016.
\newblock Equality of opportunity in supervised learning.
\newblock \emph{Advances in neural information processing systems}, 29.

\bibitem[{Keith and Stent(2019)}]{keith-stent-2019-modeling}
Keith, K.; and Stent, A. 2019.
\newblock Modeling Financial Analysts' Decision Making via the Pragmatics and Semantics of Earnings Calls.
\newblock In \emph{Proceedings of the 57th Annual Meeting of the Association for Computational Linguistics}, 493--503. Florence, Italy: Association for Computational Linguistics.

\bibitem[{Legomsky(2007)}]{legomsky2007learning}
Legomsky, S.~H. 2007.
\newblock Learning to live with unequal justice: Asylum and the limits to consistency.
\newblock \emph{Stan. L. Rev.}, 60: 413.

\bibitem[{Mehrabi et~al.(2021)Mehrabi, Morstatter, Saxena, Lerman, and Galstyan}]{mehrabi2021survey}
Mehrabi, N.; Morstatter, F.; Saxena, N.; Lerman, K.; and Galstyan, A. 2021.
\newblock A survey on bias and fairness in machine learning.
\newblock \emph{ACM computing surveys (CSUR)}, 54(6): 1--35.

\bibitem[{Morse et~al.(2021)Morse, Teodorescu, Awwad, and Kane}]{morse2021ends}
Morse, L.; Teodorescu, M. H.~M.; Awwad, Y.; and Kane, G.~C. 2021.
\newblock Do the ends justify the means? Variation in the distributive and procedural fairness of machine learning algorithms.
\newblock \emph{Journal of Business Ethics}, 1--13.

\bibitem[{Passi and Barocas(2019)}]{passi2019problem}
Passi, S.; and Barocas, S. 2019.
\newblock Problem formulation and fairness.
\newblock In \emph{Proceedings of the conference on fairness, accountability, and transparency}, 39--48.

\bibitem[{Piccolo et~al.(2023)Piccolo, Katsikouli, Gammeltoft-Hansen, and Slaats}]{piccolo2023predicting}
Piccolo, S.~A.; Katsikouli, P.; Gammeltoft-Hansen, T.; and Slaats, T. 2023.
\newblock On predicting and explaining asylum adjudication.
\newblock In \emph{Proceedings of the Nineteenth International Conference on Artificial Intelligence and Law}, 217--226.

\bibitem[{Preston(2007)}]{preston2007asylum}
Preston, J. 2007.
\newblock Big Disparities in Judging of Asylum Cases.
\newblock \emph{The New York Times}.
\newblock Accessed: 2025-01-13.

\bibitem[{Ramji-Nogales, Schoenholtz, and Schrag(2007)}]{ramji2007refugee}
Ramji-Nogales, J.; Schoenholtz, A.~I.; and Schrag, P.~G. 2007.
\newblock Refugee roulette: Disparities in asylum adjudication.
\newblock \emph{Stan. L. Rev.}, 60: 295.

\bibitem[{Rehaag(2012)}]{rehaag2012judicial}
Rehaag, S. 2012.
\newblock Judicial review of refugee determinations: The luck of the draw.
\newblock \emph{Queen's LJ}, 38: 1.

\bibitem[{Rehaag(2019)}]{rehaag2019judicial}
Rehaag, S. 2019.
\newblock Judicial review of refugee determinations (II): Revisiting the luck of the draw.
\newblock \emph{Queen's LJ}, 45: 1.

\bibitem[{Rehaag(2023)}]{rehaag2023luck}
Rehaag, S. 2023.
\newblock Luck of the Draw III: Using AI to Examine Decision-Making in Federal Court Stays of Removal.

\bibitem[{Rogers(2021)}]{rogers-2021-changing}
Rogers, A. 2021.
\newblock Changing the World by Changing the Data.
\newblock In \emph{Proceedings of the 59th Annual Meeting of the Association for Computational Linguistics and the 11th International Joint Conference on Natural Language Processing (Volume 1: Long Papers)}, 2182--2194. Online: Association for Computational Linguistics.

\bibitem[{Van~der Maaten and Hinton(2008)}]{tsnevan2008visualizing}
Van~der Maaten, L.; and Hinton, G. 2008.
\newblock Visualizing data using t-SNE.
\newblock \emph{Journal of machine learning research}, 9(11).

\bibitem[{Wang et~al.(2021)Wang, Xiao, Ma, Zhong, Tu, Zhang, Liu, and Sun}]{wang2021equality}
Wang, Y.; Xiao, C.; Ma, S.; Zhong, H.; Tu, C.; Zhang, T.; Liu, Z.; and Sun, M. 2021.
\newblock Equality before the law: Legal judgment consistency analysis for fairness.
\newblock \emph{arXiv preprint arXiv:2103.13868}.

\bibitem[{Wang, Huang, and Yao(2024)}]{wang2024procedural}
Wang, Z.; Huang, C.; and Yao, X. 2024.
\newblock Procedural fairness in machine learning.
\newblock \emph{arXiv preprint arXiv:2404.01877}.

\bibitem[{Zhao, Wang, and Derr(2023)}]{zhao2023fairness}
Zhao, Y.; Wang, Y.; and Derr, T. 2023.
\newblock Fairness and explainability: Bridging the gap towards fair model explanations.
\newblock In \emph{Proceedings of the AAAI Conference on Artificial Intelligence}, volume~37, 11363--11371.

\end{thebibliography}

\end{document}